\documentclass[journal,transmag]{elsarticle}

\usepackage{setspace}
\usepackage{amssymb,amsmath,amsthm}
\usepackage[margin=0.9in]{geometry} 

\newcommand{\bb}{\mathbf}

\usepackage[]{algorithm2e}

\begin{document}
	
\begin{frontmatter}

	\title{Clustering Gaussian Graphical Models
	}


\author[newhavenaddress]{Keith Dillon\corref{mycorrespondingauthor}}
\cortext[mycorrespondingauthor]{Corresponding author. Tel. 1-949-478-1736}
\ead{kdillon@newhaven.edu}

\address[newhavenaddress]{Department of Electrical and Computer Engineering and Computer Science, \\ University of New Haven, West Haven, CT, USA}


\begin{abstract}
We derive an efficient method to perform clustering of nodes in Gaussian graphical models directly from sample data.
Nodes are clustered based on the similarity of their network neighborhoods, with edge weights defined by partial correlations. 
In the limited-data scenario, where the covariance matrix would be rank-deficient, we are able to make use of matrix factors, and never need to estimate the actual covariance or precision matrix.
We demonstrate the method on functional MRI data from the Human Connectome Project.
A matlab implementation of the algorithm is provided.

%
%
%
%
%

\end{abstract}

\end{frontmatter}



%
%

\section{Introduction}

Gaussian graphical models \cite{whittaker_graphical_2009} are a popular and principled approach to describing networks, and are directly related to variable prediction via linear regression \cite{pourahmadi_covariance_2011}.
The focus is often on graphical model edges described by partial correlations which are zero, identifying pairs of nodes which can be treated as conditionally independent \cite{baba_partial_2004}. 
For example, the graphical LASSO \cite{friedman_sparse_2008} imposes a sparse regularization penalty on the precision matrix estimate, seeking a network which trades off predictive accuracy for sparsity.
This provides a network which interpretable and efficient to use. 
However it is not clear that sparse solutions  generalize better to new data than do dense solutions \cite{williams_back_2019}. 


Clustering of networks, on the other hand, is usually approached from a very different perspective. 
A graph is formed via some simple relationship such as affinity or univariate correlation. 
Then this network is used as a starting point for graph-theoretical approaches to partitioning.
A dominant method in this category is spectral clustering \cite{von_luxburg_tutorial_2007},
commonly described as a continuous relaxation of the normalized cut algorithm \cite{von_luxburg_tutorial_2007} for partitioning graphs.
In addition, other interpretations have been noted for spectral clustering such as random walks and smooth embeddings \cite{meila_spectral_2015}, or sometimes involving minor variants of the algorithm such as by normalizing the Laplacian differently \cite{meila_learning_2002}. 
However when the true underlying graph is based on correlations between samples of variables, the principled statistical basis for relating nodes in the network is obfuscated under multiple degrees of approximation;
first the correlation estimate is approximated with a binary (or approximately binary) edge; second the partitioning of this binary graph is approximated with a continuous relaxation. Sparse graph methods impose yet another approximation, as residual accuracy is traded off versus sparsity.

In this paper we show how clustering of partial correlations can be efficiently performed directly from the data, without need for computing the full graphical model or forming a sparse approximation. 
We provide a partial-correlation-based estimate for distance between nodes.
Then we provide an efficient algorithm for computing these distances directly from the data.
Finally we demonstrate the clustering on functional MRI data from the Human Connectome Project, a case where the network is far too large to fit in memory. 
Matlab code to implement the method is provided in an appendix.

\section{Method}

We model the data signal at the $i$th node as the zero-mean Gaussian random variable $a_i$.
The partial correlation $\rho_{i,j}$ between $a_i$ and $a_j$, is the Pearson correlation between the residuals after regressing out all other nodes except $i$ and $j$ from both.
Rather than directly performing this computationally-intensive calculation on each pair of variables using data, there are two general categories of methods used.
The first category of methods exploits the relationship with the precision matrix, the inverse of the sample covariance matrix. 
Note, of course, that if the dense network is too large to fit into memory, the so too will be the covariance and precision matrices.
The second category of methods, which we will consider here, exploits the relationship between the partial correlation and the regression coefficients for the neighborhood selection problem \cite{meinshausen_high-dimensional_2006}.
These regression coefficients are defined as the solutions $\beta_{i,j}$ to the linear system
\begin{align}
	a_i = \sum_{k \neq i} \beta_{i,k} a_k + \epsilon_i,
	\label{regression0}
\end{align}
where $a_i$ is the $i$th variable and $\epsilon_i$ is the residual.
From these $\beta_{i,j}$ we can estimate $\rho_{i,j}$ as  \cite{pourahmadi_covariance_2011},
\begin{align}
	\rho_{i,j} &= -\beta_{i,j} \sqrt{\frac{\sigma^{i,i}}{\sigma^{j,j}}}, 
	\label{rv_rho}
\end{align}
using the residual variances $\sigma^{i,i} = (Var(\epsilon_i))^{-1}$.
A common alternative formulation exploits the symmetry of the partial correlation (i.e., that $\rho_{i,j} = \rho_{i,j}$ by definition) and use the geometric mean to cancel the residual variances as in \cite{schafer_shrinkage_2005}
\begin{align}
	\rho_{i,j} &= \text{sign}(\beta_{i,j})  \sqrt{\beta_{i,j}\beta_{j,i}}.
	\label{rv_rho_geomean}
\end{align}
This also has the advantage of enforcing symmetry in sample estimates.
If the signs of  $\beta_{i,j}$ and $\beta_{j,i}$ differ, $\rho_{i,j}$ is typically set to zero.

We describe the sample version of the regression problem of Eq. (\ref{regression0}) with the linear system
\begin{align}
\bb a_i = \bb A_{-i} \boldsymbol{\beta}_i + \boldsymbol{\epsilon}_i,
\label{regression1}
\end{align}
where $\bb A_{-i}$ is the matrix $\bb A$ with the $i$th column set to zeros; the vector $\boldsymbol{\beta}_i$ is the estimates of the regression coefficients, where  $(\boldsymbol{\beta}_i)_j$ is the estimate of $\beta_{i,j}$  (defining $\beta_{i,i}=0$); and $\boldsymbol{\epsilon}_i$ is the vector of samples of the residual $\epsilon_i$.
The sample residual variances are then
\begin{align}
d_i &= 
\Vert \boldsymbol{\epsilon}_i \Vert = \Vert\bb A_{-i}^\dagger \boldsymbol{\beta}_i -  \bb a_i \Vert
\end{align} 
which we form into a vector $\bb d$ with $(\bb d)_i = d_i$.
Following \cite{dillon_computation_2019},
we write the sample version of Eq. (\ref{rv_rho}) as 
\begin{align}
\bb P = \bb D_{\bb d} \bb B \bb D_{\bb d}^{-1},
\label{P_DBD}
\end{align}
where $\bb D_{\bb d}$ is a diagonal matrix with $D_{i,i} = d_i$, and we have formed $\bb B$ with $\boldsymbol{\beta}_i$ as columns.
$\bb P$ contains our sample-based estimates of the partial correlations, with $P_{i,j}$ describing the partial correlation between nodes $i$ and $j$.
Again, we can avoid calculating the residual variances using the method of Eq. (\ref{rv_rho_geomean}) as follows,
\begin{align}
\bb P =  \text{sign}(\bb B) \odot (\bb B \odot \bb B^T)^{\circ \frac{1}{2}},
\label{P_BBB}
\end{align}
using the Hadamard product $\odot$ and element-wise exponential $\circ\frac{1}{2}$, and where the sign function is taken element-wise.
We can in turn compute $\bb B$ from the resolution matrix $\bb R = \bb A_\lambda^\dagger \bb A$ using a regularized generalized inverse $\bb A_\lambda^\dagger$, giving (\cite{dillon_computation_2019}),
\begin{align}
\bb B =\bb R_{-\bb d} \bb D_{\bb s},
\label{B_RD}
\end{align}
where $\bb D_{\bb s}$ is a diagonal matrix with $s_i = (1-R_{i,i})^{-1}$ on the diagonal, and $\bb R_{-\bb d}$ is $\bb R$ with its diagonal set to zero.
Then, combining Eqs. (\ref{P_DBD}) and (\ref{B_RD}) we get
\begin{align}
\bb P = \bb D_{\bb d} \bb R_{-d} \bb D_s \bb D_{\bb d}^{-1}.
\label{P_DBD}
\end{align}
In cases where $\bb P$ and $\bb R$ are too large to fit in memory, we can compute columns as needed,
\begin{align}
\bb p_i &= \frac{1}{d_i (1-R_{i,i})}\bb d \odot \bb r_{-i} \\
&= \bb d \odot \left(\bb A^\dagger \alpha_i \bb a_{i} - R_{i,i}\alpha_i \bb e_i\right),
\label{eq_pi_assym}
\end{align}
where we have defined $\alpha_i = [d_i (1-R_{i,i})]^{-1}$
and $\bb e_{i}$ as the $i$th column of the identity.
This requires that we calculate and store a regularized pseudoinverse of our data matrix, which is of the same size as our original data matrix.
Additionally we can pre-compute the diagonal of $\bb R$ and the $\bb d$ vector.
The diagonal of $\bb R$ is equal to the sum of its eigenvectors squared, so can be computed as a by-product of truncated-SVD regularization.
In extremely-large-scale situations, the diagonal
can be computed using even more efficient techniques 
\cite{maccarthy_efficient_2011,trampert_resolution_2013}.
The $\bb d$ vector can then be computed via
\begin{align}
d_i &= \Vert \bb A \bb B - \bb a_i \Vert \\
&= \left\vert\tfrac{1}{1-R_{i,i}}\right\vert \left\Vert \bb A \left( \bb A^\dagger \bb a_{i} - \bb e_i\right)\right\Vert. 
\label{eq_di}
\end{align}

Alternatively, combining Eqs. (\ref{P_BBB}) and (\ref{B_RD}) we get the geometric mean formulation,
\begin{align}
\bb P 
&=  \text{sign}(\bb 1 \bb s^T) \odot (\bb s \bb s^T)^{\circ \frac{1}{2}} \odot \bb R_{-d}. 
\end{align}
where $\bb s$ is the vector with elements $s_i$.
To enforce the sign test, we set $P_{i,j}$ equal to zero when $\text{sign}(s_i) \ne \text{sign}(s_j)$.
For this version of the estimate we compute columns on the fly as,
\begin{align}
\bb p_i &=  \text{sign}(s_i)  (s_i \bb s)^{\circ \frac{1}{2}} \odot \bb r_{-i} \\
&=    \vert\bb s\vert^{\circ \frac{1}{2}} \odot (\bb A^\dagger \sigma_i \bb a_i - R_{i,i}\sigma_i\bb e_i),
\label{eq_pi_sym}
\end{align}
where we have defined  $\vert\bb s\vert$ as the vector with elements $\vert s_i\vert$, and 
$\sigma_i = \text{sign}(s_i) \vert s_i\vert^\frac{1}{2} 
= \text{sign}(1-R_{i,i}) \vert 1-R_{i,i}\vert^{-\frac{1}{2}}$.
An advantage of this version, in addition to enforcing symmetry, is that the scaling only requires the diagonal of $\bb R$, not the residual $\bb d$. 
A disadvantage is potentially the need to enforce a sign test.
Generally we can write either Eq. (\ref{eq_pi_sym}) or Eq. (\ref{eq_pi_assym}) as
\begin{align}
\bb p_i &= \bb z \odot (\bb A^\dagger \zeta_i \bb a_i - R_{i,i}\zeta_i\bb e_i),
\end{align}
for appropriate definitions of $\bb z$ and $\boldsymbol\zeta$.

\subsection{Distance Calculation}

Now we will demonstrate how to compute distances using partial correlation estimates.
%
%
%
In \cite{dillon_regularized_2017} we used distances between pairs of columns of $\bb R$ as a form of connectivity-based distance in a network clustering algorithm, and demonstrated how the distance could be computed efficiently even for large datasets where $\bb R$ could not fit in memory. 
The basic idea was to use the factors $\bb A^\dagger$ and $\bb A$ in calculations, rather than precomputing $\bb R$ itself, as follows,
\begin{align}
	D^{(R)}_{i,j} &= \Vert \bb r_i - \bb r_j \Vert \\
	&= \left\Vert \bb A^\dagger\left(  \bb a_i - \bb a_j \right) \right\Vert
	\label{eq_R_dist}
\end{align}
In \cite{dillon_spectral_2018} it was noted that if SVD-truncation was used as the regularization method, then the resolution matrix can be written as
\begin{align}
	\bb R &= (\bb V \bb S_t^T \bb U^T) \bb U \bb S \bb V^T \\
	&= \bb V \bb I_r \bb V^T, 
\end{align}
where $\bb I_r$ is a truncated version of the identity, with zeros for the columns corresponding to discarded singular values.
Then the resolution distance can be written as 
\begin{align}
	D^{(R)}_{i,j} &= \Vert \bb V \bb I_r \bb v^{(i)} -  \bb V \bb I_r \bb v^{(j)} \Vert \\
	&= \left\Vert \bb I_r \left(  \bb v^{(i)} - \bb v^{(j)} \right) \right\Vert
	\label{eq_R_dist}
\end{align}
where $\bb v^{(i)}$ is the $i$th row of $\bb V$. 
So $\bb I_r \bb v^{(i)}$ is the $i$th row of the matrix formed by the first $r$ singular vectors of $\bb A$.
As these singular vectors of are the same as the singular vectors of the covariance matrix, this distance is essentially a spectral embedding of the graph formed by correlations between samples. 

We can similarly define conditional forms of distances between columns of $\bb P$, which may be implemented as
\begin{align}
	D_{i,j} &= \Vert \bb p_i - \bb p_j \Vert \\
	&= \left\Vert \bb z \odot\bb A^\dagger\left( \zeta_i \bb a_i - \zeta_j\bb a_j \right) 
	-(R_{i,i} z_i \zeta_i \bb e_i 
	-R_{j,j} z_j \zeta_j \bb e_j) 
	\right\Vert
	\label{eq_P_dist}
\end{align}
wherein we can see the similarity to the resolution form of the spectral clustering distance in Eq. (\ref{eq_R_dist}).

\subsection{Partial correlation clustering}

Next we demonstrate the use of the partial correlation distance in clustering, by similarly extending $k$-means clustering.
A basic $k$-means clustering algorithm for clustering the columns of some matrix $\bb M$ is given in Algorithm 1.
\begin{algorithm}
	1. Choose number of clusters $K$ and initialize cluster centers $\bb c_k, \; k=1,...,K$\;
	\While{Convergence criterion not met}{
		2. Calculate distances $D_{ik}$ between every column $\bb m_k$ and every cluster center $\bb c_i$\;
		3. Label each column as belonging to nearest cluster center: $l_k = \arg \min_i D_{ik}$\;
		4. Recalculate cluster centers as mean over data columns with same label: $\bb c_i = \frac{1}{\vert S_i \vert}\sum_{j \in S_i} \bb a_j$,  where $S_i = \left\{k | l_k = i \right\} $ \;
	}
	\caption{Basic $k$-means applied to data columns of $\bb M$.}
\end{algorithm}

For a small dataset we may form the matrix $\bb P$ and directly apply this algorithm to its columns.
However, for large datasets we cannot fit $\bb P$ in memory so must maintain factors as in Eq. (\ref{eq_P_dist})
In particular, the squared distances between a given center $\bb c_i$ and a column $\bb p_k$ of $\bb P$ can be calculated as
\begin{align}
D_{ik}^2 &= \Vert \bb c_i - \bb p_k \Vert_2^2 \notag \\
&= \bb c_i^T \bb c_i + \bb p_k^T \bb p_k - 2 \bb c_i^T \bb p_k.
\end{align}
Since we are only concerned with the class index $i$ of the cluster with the minimum distance to each column, we do not need to compute the $\bb p_k^T \bb p_k$ term, so we can compute the labels as
\begin{align}
l_k &= \arg \min_i D_{ik}^2 \notag \\
&= \arg \min_i \left\{ \bb c_i^T\bb c_i -  2 (\bb c_i \odot \bb z)^T (\bb A^\dagger \zeta_k \bb a_k - R_{k,k}\zeta_k\bb e_k) \right\}.
\end{align}
By forming a matrix $\bb C_{\bb z}$ with weighted cluster centers $\bb c_i \odot \bb z$ as columns, and a weighted data matrix $\bb A_\zeta$ with $\zeta_i\bb a_i$ as columns, we can efficiently compute the first part of the cross term for all $i$ and $k$ as 
$(\bb C_{\bb z}^T \bb A^\dagger) \bb A_\zeta$, 
a $K$ by $n$ matrix.
The second part of the cross term can be computed by (element-wise) multiplying each row of $\bb C_{\bb z}^T$ by a vector who's $k$th element is $R_{k,k}\zeta_k$. 
%
%
%
%
%

Similar tactics can be used to efficiently compute the mean over columns in each cluster, by noting that the mean over a set of columns with $S$ as the set of column indices can be written as
\begin{align}
\bb c_i &= \frac{1}{\vert S \vert}\sum_{j \in S} \bb p_j \notag \\
&= \frac{1}{\vert S \vert}\bb z \odot \left( \bb A^\dagger \sum_{j \in S}  \zeta_j \bb a_j 
- \sum_{j \in S} R_{j,j}\zeta_j\bb e_j
\right) 
\end{align}
So in general, we see that clustering of $\bb P$  can be implemented whenever the dataset itself is small enough to implement $k$-means clustering, taking roughly double the storage space and computational resources (rather than the number of variables squared).
In the Appendix we provide matlab \cite{matlab_version_2010} code for an efficient implementation of the clustering algorithm, taking advantage of fast broadcast routines for matrix computations.

\section{Results}

\begin{figure}[h!] \centering 
	\scalebox{0.4}{\includegraphics[clip=true, trim=0in 0in 0in 0in]{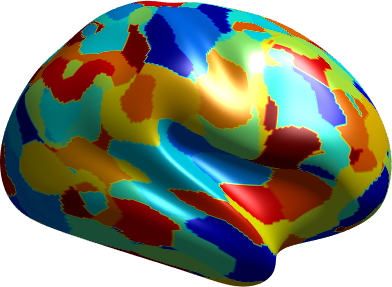}} 
	\scalebox{0.4}{\includegraphics[clip=true, trim=0in 0in 0in 0in]{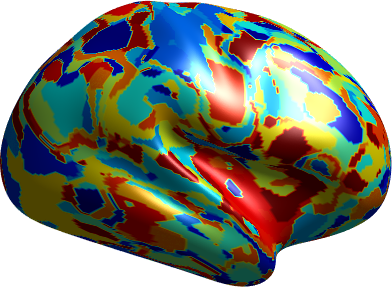}} 
	\scalebox{0.4}{\includegraphics[clip=true, trim=0in 0in 0in 0in]{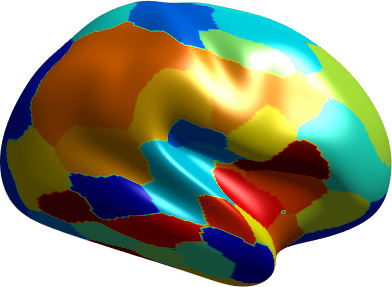}} 
	\caption{Clustering functional MRI data for single subject from the HCP project into 100 clusters; data contains 96854 time series with 1200 time samples each; direct clustering of time series (left); clustering of correlations between time series (middle); clustering of partial correlations between time-series (right).}
	\label{fig_hcp}
\end{figure}

Fig. \ref{fig_hcp} gives a demonstration of the algorithm applied to a functional Magnetic Resonance Imaging (fMRI) dataset for a subject from the Human Connectome Project \cite{van_essen_human_2012}, compared to other clustering approaches.
The data was preprocessed by applying spatial smoothing with a 5mm kernel, and SVD-truncation-based regularization to achieve a cutoff of 30 percent of singular values. 
The data contains 96854 time series with 1200 time samples each, resulting in a data matrix $\bb A$ of size $1200 \times 96854$. 
Each time-series describes the blood-oxygen-level dependent (BOLD) activity in one voxel of the brain, so a network formed by comparing these signals provides an estimate of the functional connectivity of the brain.
A clustering of this dense network would produce clusters of nodes with similar connectivity, estimating the modularity of function in the brain.
The dense network describing the relationships between all pairs of voxels, however, would require a dense $\bb P$ matrix of size $96854\times 96854$, which is far too large to fit in computer memory. 
However the limited rank of this matrix means we only need store the  $96854\times 1200$  pseudoinverse. The clustering algorithm took 9 seconds on a desktop computer.
We see that clustering of $\bb P$ produces much more modular segmentation of the regions of the brain, particularly compared to the conceptually-similar approach of clustering the network of univariate correlations of the data instead.

\begin{figure}[h!] \centering 
	\scalebox{0.7}{\includegraphics[clip=true, trim=0in 0in 0in 0in]{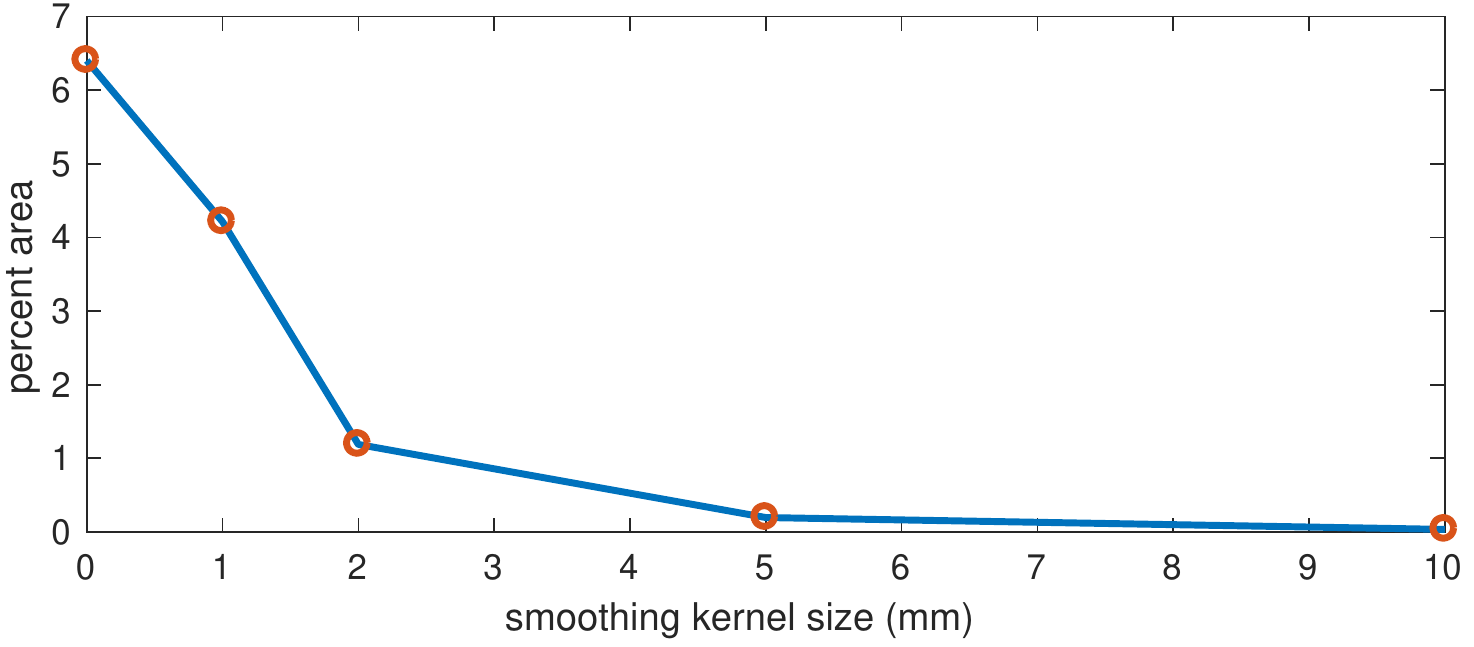}} 	
	\caption{Plot of difference between R and P method.}		
	\label{fig_R_vs_P_plot}
\end{figure}

Next we considered the effect of the difference between Eqs. (\ref{eq_R_dist}) and (\ref{eq_P_dist}) for this dataset. 
In Fig. \ref{fig_R_vs_P_plot} we plot the percentage of nodes which fall into different clusters using the two different distance calculations, as increasing amounts of spatial smoothing are applied. 
Spatial smoothing creates correlations between neighboring variables.
Therefore, the removal of the $i$th and $j$th variables will have less effect, as for example, the same information is increasingly included in the $(i+1)$th and $(j+1)$th variables due to smoothing.
Interestingly, we see that even with no spatial smoothing applied, clustering of partial correlations for this data agrees within 94 percent of the simpler spectral estimate.

\section{Discussion}

We showed how partial correlations can be clustered by providing an appropriate correction to a simple spectral distance calculation. 
We also showed that this could be computed efficiently by utilizing the data matrix and its pseudoinverse. 
As the latter may be computed by computing singular vectors of the data matrix, the proposed method can be performed with comparable efficiency to spectral clustering.  
We also demonstrated the approach on a functional brain imaging dataset, where we found a strong agreement with a spectral distance calculation. 
This suggests another perspective on the success of spectral clustering methods--as an approximation to clustering of a Gaussian graphical model for the data. 
A benefit of the proposed approach and perspective is its principled basis; 
rather than perform a series of approximations to convert a covariance selection problem into a intractable graph parcellation problem which must be relaxed, we can directly cluster variables based on their partial correlations.

The drawbacks of the proposed approach include the moderately increased computational effort, requiring double the storage, and approximately two matrix-vector products instead of one. Still, this is only a factor of two. 
The handling of negative values is another difficulty of the method, as will all similar statistical methods. 
Though our approach to address it has the advantage of leveraging an approach used in bioinformatic analyses for several years.


%
%
%
%
%

Finally, there are a number of potential extensions and directions for research based on this result. 
Instead of a simple $k$-means stage, we might apply a more sophisticated clustering algorithm such as fuzzy or hierarchical clustering.
Given the principled statistical basis for our distance calculation, we might also utilize more sophisticated statistical metrics.
For example, the penalized regression may be viewed as a maximum a posteriori estimate; we might extend this idea to compute a kind of Bayesian clustering based on computing moments of the distribution of each variable, rather than a point estimate.

\section{Appendix: matlab implementation of clustering}

In this appendix we demonstrate how the clustering of a dense $\bb P$ matrix can be performed efficiently when the data matrix is of limited rank.
We make use of the more efficient form of the regularized pseudoinverse for an underdetermined matrix, as well as efficient broadcast methods where possible. 

\begin{verbatim}
A = randn(500,100000); % data matrix
lambda = 1; % regularization parameter
k = 100; % number of clusters

[rows_A,cols_A] = size(A);

% standardize data columns
A = bsxfun(@minus,A,mean(A));
A = bsxfun(@times,A,1./sum(A.^2).^.5); 

% compute diagonal of R via sum of squared eigenvectors
[uA,sA,vA] = svd(A,'econ'); 
r = sum(vA(:,1:rank(A)).^2,2)';

% compute pseudoinverse efficiently
iA_lambda = A'*inv(A*A'-lambda*eye(rows_A));

% compute scaling vectors (symmetric version)
s = 1./(1-r(:));
z = abs(s(:)).^.5;
zeta = sign(s).*abs(s(:)).^.5; 

Az = bsxfun(@times,A,z(:)');

% randomly assign columns to clusters initially
c = ceil(rand(cols_A,1)*k); 

n_change = inf
while (n_change>0)
  M = sparse(1:cols_A,c,1,cols_A,k,cols_A); % cols of M are masks of clusters
  M = bsxfun(@times, M, 1./sum(M,1));  % now M is averaging operator

  P_c_1 = iA_lambda*(Az*M);  % first part of cluster center calc
  P_c_2 = bsxfun(@times,M,r.*zeta); % second park (peak removal)
  P_c = bsxfun(@times,A_c_1-A_c_2,z(:)); % cluster centers

  Pz2_c = sum(P_c.^2,1); % squared term from distance

  Cz = bsxfun(@times,P_c,z(:)); % weighted cluster centers
  D_ct1 = (Cz'*iA_lambda)*Az; % first part of cross-term
  D_ct2 = bsxfun(@times,Cz',r'.*zeta(:)'); % second part of cross term
  D_ct = D_ct1-D_ct2; % cross-term
  
  Dz = bsxfun(@minus,D_ct,.5*Pz2_c'); % dist metric (sans unnecessary term) 
  
  c_old = c;
  [D_max,c(:)] = max(Dz,[],1); % c is arg of max

  n_change = sum(c~=c_old);
  disp(n_change);
end;

\end{verbatim}

\section*{References}
%

\end{document}